\documentclass[a4paper,10pt,twocolumn]{article}

\usepackage[utf8]{inputenc}
\usepackage[english]{babel} 
\usepackage{vmargin}
\usepackage{authblk}
\usepackage{subcaption}
\usepackage{multirow}
\usepackage{pbox}
\usepackage{graphics}

\usepackage{amssymb}
\usepackage{graphicx}			
\usepackage{amsmath}			

\setpapersize{A4}
\setmargins{2.5cm} 
{1.5cm}            
{16.5cm}           
{23.42cm}          
{10pt}             
{1cm}              
{0pt}              
{2cm}              

\title{Zorro: A Flexible and Differentiable Parametric Family of Activation Functions That Extends ReLU and GELU}
\author[1]{Matías Roodschild}
\author[1]{Jorge Gotay-Sardiñas}
\author[1]{Victor A. Jimenez}
\author[1]{Adrián Will}
\affil[1]{\small Grupo de Investigación en Tecnologías Informáticas Avanzadas (GITIA), Universidad Tecnológica Nacional - Facultad Regional Tucumán}

\date{}

\usepackage{fancyhdr}

\hypersetup{pdfborder = {0 0 0}}

\begin{document}

\maketitle

\pagestyle{fancy}
\thispagestyle{fancy}

\footnotetext[1]{Grupo de Investigación en Tecnologías Informáticas Avanzadas (GITIA), Universidad Tecnológica Nacional, Facultad Regional Tucumán, Rivadavia 1050 (4000), San Miguel de Tucumán, Tucumán, Argentina.\\
\indent E-mails: mroodschild@gmail.com, jgotay57@gmail.com, victoradrian.jimenez@frt.utn.edu.ar (corresponding author), adrian.will@gitia.org.}
\begin{abstract}
\small
Even in recent neural network architectures such as Transformers and Extended LSTM (xLSTM), and traditional ones like Convolutional Neural Networks, Activation Functions are an integral part of nearly all neural networks. They enable more effective training and capture nonlinear data patterns. More than 400 functions have been proposed over the last 30 years, including fixed or trainable parameters, but only a few are widely used. ReLU is one of the most frequently used, with GELU and Swish variants increasingly appearing. However, ReLU presents non-differentiable points and exploding gradient issues, while testing different parameters of GELU and Swish variants produces varying results, needing more parameters to adapt to datasets and architectures. This article introduces a novel set of activation functions called Zorro, a continuously differentiable and flexible family comprising five main functions fusing ReLU and Sigmoid. Zorro functions are smooth and adaptable, and serve as information gates, aligning with ReLU in the 0-1 range, offering an alternative to ReLU without the need for normalization, neuron death, or gradient explosions. Zorro also approximates functions like Swish, GELU, and DGELU, providing parameters to adjust to different datasets and architectures. We tested it on fully connected, convolutional, and transformer architectures to demonstrate its effectiveness.

\end{abstract}

\section{Introduction}
\label{sec:intro}

Although some variations of Transformers and recent architectures have no explicit activation functions, they are still an integral part of most neural network architectures. Although only a handful are effectively used in practice, over 400 Neural Network activation functions have been defined over the last 30 years \cite{kunc_three_2024}. Activation functions are a simple yet effective way to allow the network to capture nonlinear patterns in the data and get prescribed behaviors like Logistic Sigmoid functions in LSTM networks, ReLU functions in Transformers, and others. Nevertheless, not every function works for every architecture or dataset, so many adaptive functions that include trainable parameters have been proposed \cite{mastromichalakis_parametric_2023,delfosse_adaptive_2024,martinez-gost_enn_2024}. These parameters allow the activation function to be adapted to the particular dataset and not only the architecture.

In that line, designing an effective activation function or even knowing which one is the most appropriate for a given architecture and dataset is still an active area of research. Only in the last six months have there been over six different proposals \cite{rajanand_erfrelu_2024,subramanian_apalu_2024,noel_significantly_2024,delfosse_adaptive_2024,sun_method_2024,martinez-gost_enn_2024}. Moreover, parametric activation functions that can adapt their behavior to the dataset during training are an effective and relatively simple way to improve the results without significantly increasing the processing time. Alternatively, the Extended LSTM (xLSTM), a recent architecture proposed by the original author of LSTM networks \cite{xLSTM_2024}, includes the Swish activation function and claims to achieve better generalization errors than current Transformer networks. Block Recurrent Transformers \cite{BlockRecTransf} is another recent architecture that uses an exponential function as an activation function. An adaptive, multi-parametric, and differentiable function that can approximate many of the usual activation functions (Swish, SiLU, GELU, ReLU, among others) should allow future researchers and developers many possibilities.

In this work, we define and study a multi-parametric family of activation functions called \textit{Zorro}, proposing five variations (Symmetric-Zorro, Asymmetric-Zorro, Sigmoid-Zorro, Tanh-Zorro, and Sloped-Zorro). All these functions were designed as a combination of ReLU and Sigmoid activation functions and were inspired by DGELU and DSiLU.
DSiLU, the derivative of the SiLU activation function, was proposed as an activation function for Convolutional Networks in \cite{elfwing_sigmoid-weighted_2017}. DGELU, the derivative of the GELU function, has not been proposed as an activation function up to date. So, Zorro has the same general shape as these functions but preserves the linear central part in $[0, 1]$ characteristic of the ReLU function. This design allows Zorro to provide similar results for the general case where data is normalized or initialized for the ReLU function, improving the results above 1 and below 0. The rest of the functions are defined by reparametrizations and rescaling so that they approximate their namesakes: Sigmoid-Zorro has a similar range and derivative in zero as the Sigmoid function, making it appropriate for use as a gating function; Tanh-Zorro has range slightly more extensive than $[-1, 1]$, is centered in zero, and with derivative 1 in zero, similar to the original Tanh function; Asymmetric-Zorro exploits the fact that different coefficient for negatives and values over one can produce better results; Finally, Sloped-Zorro increases the derivative of the linear part so that training is faster than others. The paper is completed by an in-depth test on a simple feedforward architecture that shows that, even though they are all reparametrizations of the same basic function, their behavior as activation functions is entirely different, making them suitable for different architectures and datasets.

The rest of this work is organized as follows: Section \ref{sec:related_works} shows the related works, highlighting important aspects of the preexisting activation functions; Section \ref{sec:funciones_tradicionales} describes the most commonly used activation functions; Section \ref{sec:zorro} presents the Zorro function proposed in this work, describing its characteristics and different variants; Section \ref{sec:ajuste_parametros} presents a parameter adjustment analysis using a fully connected dense feedforward network; Section \ref{sec:cnn} presents the results of applying the proposed functions in convolutional architecture on 5 different datasets; Section \ref{sec:transformer} presents the results obtained by applying variants of the Zorro function in a Transformer architecture; Finally, Section \ref{sec:conclusions} presents the general conclusion and perspectives of future works.


\section{Related Works}
\label{sec:related_works}

Activation function design is a very active area of research. 
Besides the thorough and comprehensive review \cite{kunc_three_2024}, Website Papers With Code \cite{Pag_Papers_with_Code_Activ}, a recognized website for keeping records of state-of-the-art machine learning algorithms, reports 74 of the most frequently used activation functions. Moreover,
\cite{dubey_activation_2022} conducts an extensive and systematic study of activation function from perspectives such as function shape, differentiability, boundedness, number of parameters, whether or not it is a monotone function, and computer efficiency. 

As for the frequency of use in software packages, LLMs, CNNs, and other commercially, academic, and widely used architectures, we can mention ReLU, Sigmoid, Tanh, Sigmoid-Weighted Linear Units (SiLU) \cite{elfwing_sigmoid-weighted_2017}, Gaussian Error Linear Unit (GELU) \cite{hendrycks2023gaussian}, Swish \cite{Ramachandran2017}, ELU \cite{clevert_elus_2015}, and Leaky ReLU \cite{Maas2013} (among many others). 
There also exist parametric versions of many of those, including Shifted and Scaled Sigmoid \cite{Shifted_Sigmoid_2018} and Scaled Tanh \cite{LeCun98}, which include a slope parameter, a parametric version of Leaky ReLU (PLeaky ReLU) \cite{He2015}, Parametric Leaky Tanh \cite{mastromichalakis_parametric_2023}, and more sophisticated approaches, including ErfReLU \cite{rajanand_erfrelu_2024} which is ReLU with an adaptability parameter that can be trained along with the weights. This is a partial list since, once again, there are a considerable number of functions.

Other than those frequently used activation functions that have produced good results in a wide range of applications, datasets, and architectures, some functions for particular purposes have been devised. As a small sample of the many such functions, ASU \cite{rahman_amplifying_2023} is a function designed to cope with the specific case of vibrations, Signed and Truncated Logarithm Activation Function (STL) \cite{gong_stl_2023}, based on a logarithmic function that produces an entirely different behavior, and the functions proposed in \cite{serrano_hosc_2024} were designed to preserve sharpness in signals or images. 
Finally, recent architectures like AAREN Neural Networks include exponential activations \cite{AAREN_2022}, in whole or only on the positive side \cite{biswas_non-monotonic_2023}, Extended LSTM (xLSTM) that includes ReLU and Swish \cite{xLSTM_2024}, and transformer architectures that still include ReLU activation functions. 

None of the functions collected in those reviews and recent publications \cite{subramanian_apalu_2024,noel_significantly_2024,sun_method_2024} coincide with the ones proposed in this work. The closest are DSiLU and DGELU, as depicted in the next section. Moreover, DSiLU was proposed as an activation function for a Convolutional architecture \cite{elfwing_sigmoid-weighted_2017}. However, as far as we have checked, DGELU has never been proposed or analyzed as an activation in the literature.


\section{Some relevant activation functions}
\label{sec:funciones_tradicionales}

\subsection{Generalized Sigmoid (GSigmoid)}
\label{sec:sigmoid}

\begin{figure}[htb]
    \centering
    \includegraphics[width=\columnwidth]{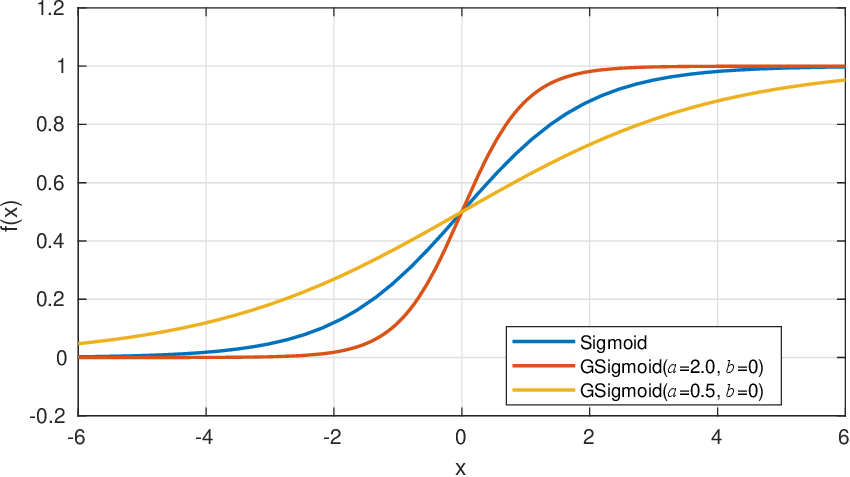}
    \caption{Logistic Sigmoid and GSigmoid for different values of the $a$ parameter.}
    \label{fig:gsigmoid}
\end{figure}

The Logistic Sigmoid is one of the first functions for neuron activation, and it is widely used in several architectures. 
It is defined by Equation \ref{eq:sigmoid}. The Shift and Scaled Sigmoid \cite{Shifted_Sigmoid_2018} is a more generalized version of this function, which includes two parameters to provide flexibility. For simplicity, in this work, we call this function Generalized Sigmoid ($GS$), defined by Equation \ref{eq:gsigmoid}, where the parameter $a$ controls the increase in slope, and $b$ is a horizontal shift. Figure \ref{fig:gsigmoid} shows the curve obtained with different values of $a$, considering $b$ equal to zero for clarity.
\begin{equation}
    \sigma(x)=\frac{1}{1+e^{-x}}
    \label{eq:sigmoid}
\end{equation}
\begin{equation}
    \text{GS}(x, a, b) = \sigma(a(x-b))  
    \label{eq:gsigmoid}
\end{equation}

This function preserves the differentiability and boundedness properties of the original Sigmoid function. For $a > 1$, it increases the maximum value for its derivative, acquiring benefits in some neural network training contexts. The reparametrization of the Logistic Sigmoid function allows us to define the functions proposed in this work easily.



\subsection{ReLU-based functions}
\label{sec:relu}

Rectified Linear Unit (ReLU) is one of the most widely used activation functions in many neural network architectures. Its linear part with the derivative equal to 1 for positive input values, simplicity of computation, and efficiency make it the preferred choice in many problems. Mathematically, it is defined by Equation \ref{eq:relu}. 
\begin{equation}
    \text{ReLU}(x)=\text{max}(0,x)
    \label{eq:relu}
\end{equation}

\begin{figure}[htb]
    \centering
    \includegraphics[width=\columnwidth]{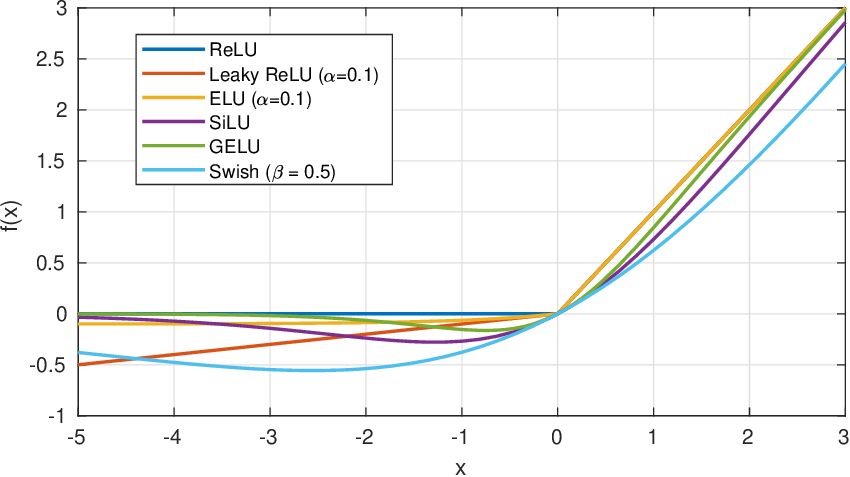}
    \caption{Comparison of ReLU-based, ELU and Swish-based activation functions.}
    \label{fig:activaciones_relu}
\end{figure}

Despite having many advantages, this function causes the Explosive Gradient Problem (EGP) \cite{Philipp2017}, where the gradient becomes unstable and usually must be clipped (gradient clipping). In addition, it converts all negative inputs to zero, producing a large area with zero derivatives. For this reason, some neurons stop updating their weights during training (neuron death). Variants of the ReLU function, such as the Leaky Rectified Linear Unit (Leaky ReLU) \cite{Maas2013}, were proposed to mitigate this problem. The Leaky ReLU replaces the zero value for negative inputs with a small fraction of the input ($\alpha x$ where $\alpha$ is a small value, usually 0.01). ReLU and Leaky ReLU are shown in Figure \ref{fig:activaciones_relu}, along with other activation functions.

\subsection{Exponential Linear Unit (ELU)}
\label{sec:elu}

The Exponential Linear Unit (ELU) \cite{Clevert2015} is an alternative to the ReLU function, designed to reduce the Vanishing Gradient Problem (VGP) \cite{hochreiter1998vanishing}. The ELU function is defined by Equation \ref{eq:elu}, where the hyperparameter $\alpha$ controls the saturation rate for negative inputs. It is the identity for positive inputs and presents negative values for $x<0$, being differentiable in the whole domain (see Figure \ref{fig:activaciones_relu}). The parameter can be set to have the mean of activation values closer to zero, avoiding problems of bias shifting between layers during training and achieving faster and more stable learning than some of its counterparts.
\begin{equation}
    \text{ELU}(x, \alpha) = 
    \begin{cases}
        \alpha (e^x-1) & x \leq 0 \\
        x              & x > 0 \\
    \end{cases}
    \label{eq:elu}
\end{equation}


\subsection{Swish-based functions (SiLU and GELU)}
\label{sec:swish}

The Swish function \cite{Ramachandran2017} was designed as a smooth replacement for the ReLU as an activation function. Equation \ref{eq:swish} defines it by multiplying $x$ and the Sigmoid function, and the $\beta$ parameter controls the shape of the function around zero. It is unbounded for positive values and tends to zero for negative values. 
\begin{equation}
    \text{Swish}(x, \beta)=x \sigma(\beta x) 
    \label{eq:swish}
\end{equation}

Different activation functions represent particular cases of the Swish function. One is the Sigmoid-Weighted Linear Units (SiLU) \cite{elfwing_sigmoid-weighted_2017} defined by Equation \ref{eq:silu} by multiplying $x$ and the Sigmoid function. So, it is obtained from the Swish function by setting 1 for the $\beta$ parameter. 
\begin{equation}
    \text{SiLU}(x)=x \sigma(x)
    \label{eq:silu}
\end{equation}

A more probabilistic approach for the same problem of finding a differentiable yet efficient replacement for the ReLU function produced the Gaussian Error Linear Unit (GELU) \cite{hendrycks2023gaussian}. Recently, GELU has been increasingly used in transformer architectures \cite{lee_gelu_2023}. It is defined by Equation \ref{eq:gelu}, where $\sigma$ is the Logistic Sigmoid function, and $erf$ is the Gauss Error Function. Since it is complex and expensive to calculate, the original authors suggested and analyzed an approximation using the Swish function with $\beta = 1.702$. So, this is another particular case of the Swish function. Figure \ref{fig:activaciones_relu} compares ReLU-based functions, the ELU function, and Swish-based functions.
\begin{equation}
    \text{GELU}(x) = \frac{x}{2} \left[1+erf\left(\frac{x}{\sqrt{2}}\right)\right] \approx x \sigma(1.702 x)
    \label{eq:gelu}
\end{equation}


\subsection{Derivatives of the SiLU and GELU functions}
\label{sec:Dsilu_Dgelu}

\begin{figure}[htb]
    \centering
    \includegraphics[width=\columnwidth]{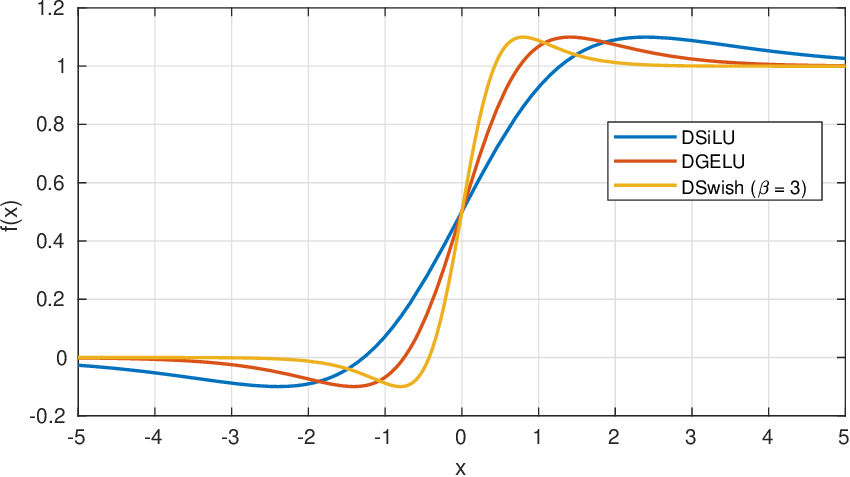}
    \caption{Comparison of DSwish-based activation functions.}
    \label{fig:dswish}
\end{figure}

The derivative of the function Swish is defined by Equation \ref{eq:dswish} and shown in Figure \ref{fig:dswish}. For $\beta=1$, it is the derivative of the function SiLU (DSiLU); for $\beta = 1.702$, it is the derivative of the function GELU (DGELU). It is clear that they are reparametrizations of each other and have the same general shape. 
\begin{equation}
    \text{DSwish}(x, \beta) =\beta \text{Swish}(x, \beta) \left[ 1 - \sigma(\beta x)\right] + \sigma(\beta x)
    \label{eq:dswish}
\end{equation}

Now being bounded and differentiable among other relevant properties, they could be proposed as activation functions in their own right. DSiLU has been proposed as an activation function for Convolutional Networks \cite{elfwing_sigmoid-weighted_2017}. Still, we have not found works that propose DGELU as an activation function in the literature. They have a similar shape as the functions proposed in this work, so they will be compared to determine which provides better results.


\section{A new activation function: Zorro}
\label{sec:zorro}

ReLU is a very powerful but simple activation function. Its definition allows for a straightforward and fast implementation, and a simple normalization or a smart initialization usually fixes any problem due to the non-differentiability around zero. However, the tendency to generate explosive gradients for input values greater than one makes it very complicated for the training algorithm to fit the data in these zones correctly. This problem is critical in industrial environments, where a slight improvement in accuracy can be costly because it comes from a small percentage of the data in those zones.

On the other hand, DSiLU and DGELU are bounded functions, differentiable, not excessively complex to calculate, and have an almost linear part in the central critical zone. These characteristics make them worthy of being considered an activation function. The DSiLU function was the first to be proposed as an activation function for convolutional networks, producing good results \cite{elfwing_sigmoid-weighted_2017}. We have not found in the literature that DGELU has been considered as an activation function.

In order to leverage the advantages of ReLU and solve its drawbacks, we combine it with the Sigmoid function in a shape similar to the DGELU function but preserving the linear part of the ReLU in $[0, 1]$. This combination gives rise to the \textit{Zorro} function defined by Equation \ref{ec:zorro}, where $GS$ is the Generalized Sigmoid function, $a$ and $b$ are fixed parameters, and $k$ is a coefficient defined by Equation \ref{ec:constante_zorro}. The fixed parameters control the shape and position of the characteristic ``humps'' similar to those of the DGELU function, and the $k$ coefficient ensures the differentiability of the function at $x = 0$ and $x = 1$. We use the GSigmoid function for the Zorro definition to simplify its mathematical expression. Also, many software packages have optimized versions of the Sigmoid function that would reduce computations and speed up training.

\begin{equation} %
    \newlength\eqmaxwidthz
    \setlength{\eqmaxwidthz}{\columnwidth-22pt}
    \resizebox{\ifnum\width<\eqmaxwidthz \width \else \eqmaxwidthz\fi}{!}{$%
        \text{Zorro}_\text{sym}(x, a, b) = 
        \begin{cases}
            k x \text{GS}(x, a, b)                          & \hskip -5pt x < 0 \\
            x                                               & \hskip -5pt x\in[0,1] \\ 
            1 \text{-} k (1\text{-}x) \text{GS}(1\text{-}x, a, b)  & \hskip -5pt x > 1
        \end{cases}
    $}
    \label{ec:zorro}
\end{equation}
\begin{equation}
    k = 1 + e^{ab} 
    \label{ec:constante_zorro} 
\end{equation}

\begin{figure}[htb]
    \centering
    \includegraphics[width=\columnwidth]{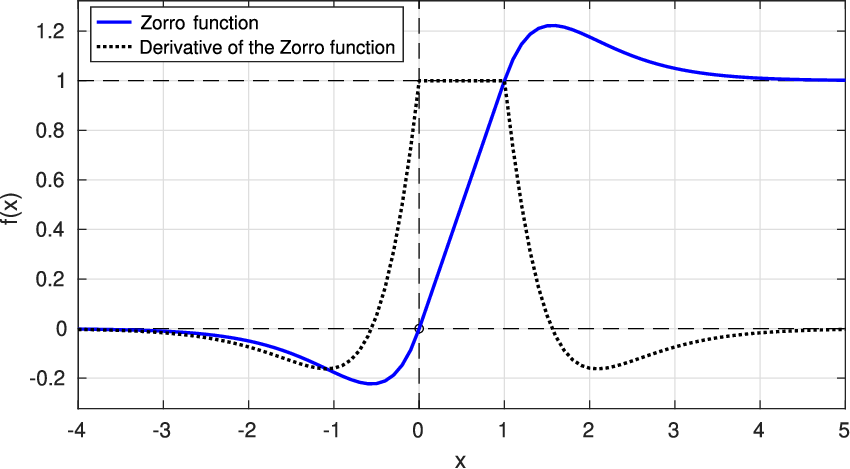}
    \caption{The Zorro Activation Function and its derivative ($a=2$ and $b=0.5$).}
    \label{fig:zorro}
\end{figure}

Figure \ref{fig:zorro} shows the Zorro function for particular values of $a$ and $b$, where we can see how it preserves the central linear part, is similar to SiLU for $x<0$, and is reflected around the point $(0.5, 0.5)$ for $x>1$ to preserve the symmetries like DSiLU or DGELU functions. Larger values of the parameter $a$ take the humps closer to the asymptotic horizontal lines $y=0$ and $y=1$, while larger values of $b$ move the maximum and minimum values located in the humps away from zero. 

The formula for the derivative can be seen in Equation \ref{eq:derivada_zorro}. It confirms that Zorro is differentiable, although the second derivative does not exist in $0$ and $1$. A somewhat more involved calculation shows that it is bounded for $a > 0$ since for $a=0$ and $b=0$, it is the identity. 
That is, it can be used as an activation function and trained using backpropagation.
\begin{equation}
    \newlength\eqmaxwidthdz
    \setlength{\eqmaxwidthdz}{\columnwidth-27pt}
    \resizebox{\ifnum\width<\eqmaxwidthdz \width \else \eqmaxwidthdz\fi}{!}{$%
        \frac{d}{dx}\text{Zorro}_\text{sym}(x,a,b) =
        \begin{cases}
            \begin{aligned}
                & k\text{GS}(x,a,b) [1 - \\
                & \quad a x(1 \text{-} \text{GS}(x,a,b))] \\
            \end{aligned} & \hskip -6pt x<0\\
            1             & \hskip -6pt x\in[0,1] \\ 
            \begin{aligned}
                & k\text{GS}(1\text{-}x,a,b) [1 -  \\
                & \quad a (1\text{-}x)(1 \text{-} \text{GS}(1\text{-}x, a, b))] \\
            \end{aligned} & \hskip -6pt x > 1\\
        \end{cases}
    $}
    \label{eq:derivada_zorro}
\end{equation}


\subsection{Variants of the Zorro Activation Function}
\label{sec:zorro_variantes}

Many variants can be defined based on the Zorro function described before. The following are the most interesting ones that produce good results in practice.


\subsubsection{Asymmetric-Zorro}
\begin{figure}[htb]
    \centering
    \begin{subfigure}{0.493\columnwidth}
        \centering
        \includegraphics[height=0.82\textwidth]{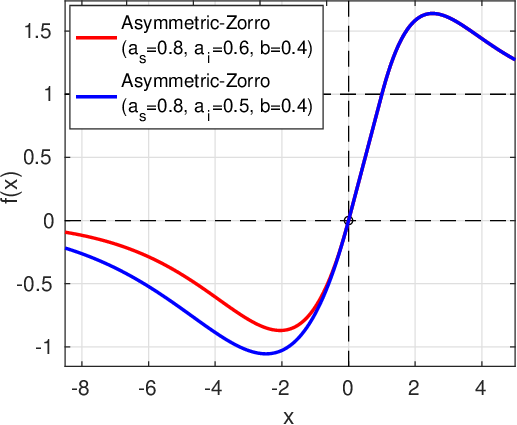}
        \caption{Asymmetric-Zorro.}
        \label{fig:zorro_asimetrica}
    \end{subfigure}
    \begin{subfigure}{0.493\columnwidth}
        \centering
        \includegraphics[height=0.82\textwidth]{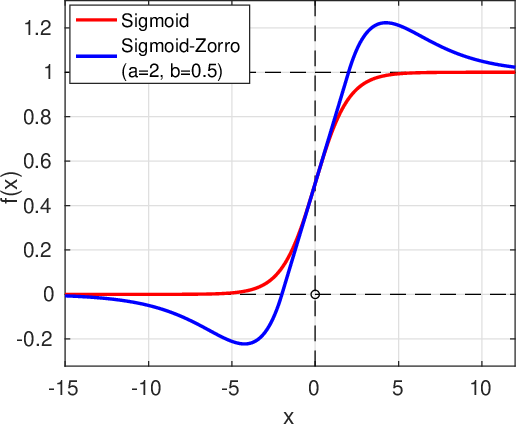}
        \caption{Sigmoid-Zorro.}
        \label{fig:zorro_sigmoid}
    \end{subfigure}
    \begin{subfigure}{0.493\columnwidth}
        \centering
        \includegraphics[height=0.82\textwidth]{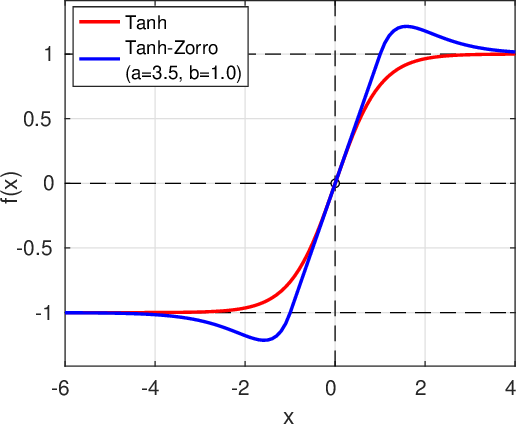}
        \caption{Tanh-Zorro.}
        \label{fig:zorro_tanh}
    \end{subfigure}
    \begin{subfigure}{0.493\columnwidth}
        \centering
        \includegraphics[height=0.82\textwidth]{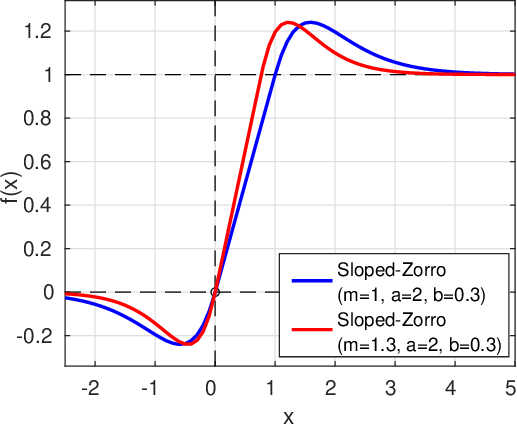}
        \caption{Sloped-Zorro.}
        \label{fig:zorro_sloped}
    \end{subfigure}
    \caption{Most relevant variants of the Zorro family activation function.}
    \label{fig:zorro_variants}
\end{figure}

The mean value of the activation function over the entire dataset is a good indicator of convergence: It has been shown that if this mean value is not zero, it acts as a bias for the next layer, delaying the training process \cite{Clevert2015}. Thus, functions with zero mean, such as Tanh, will be more effective than Sigmoid, and functions with lower mean in a given interval symmetric around zero, such as SiLU, GELU, and Swish, will perform better than ReLU.  So, an asymmetric version of Zorro with a smaller mean value will have these benefits and be able to train deeper networks. The asymmetric variant of the Zorro function arises by considering independent values of the coefficients $a$ and $b$ for the positive and negative parts, allowing asymmetric humps. The Asymmetric-Zorro function is then defined by Equation \ref{eq:zorro_asym}. It is shown in Figure \ref{fig:zorro_asimetrica} for different parameter values.
\begin{equation}
    \newlength\eqmaxwidthza
    \setlength{\eqmaxwidthza}{\columnwidth-27pt}
    \resizebox{\ifnum\width<\eqmaxwidthza \width \else \eqmaxwidthza\fi}{!}{$%
        \text{Zorro}_\text{asym}(x, a_{s}, a_{i},  b) =
        \begin{cases}
            k_{i} x  \text{GS}(x, a_{i},  b) & \hskip -5pt  x<0 \\ 
            x                                & \hskip -5pt  x\in[0,1] \\ 
            1 \text{-} k_{s} (1 \text{-} x) \text{GS}(1 \text{-} x, a_{s}, b) & \hskip -5pt  x>1
        \end{cases}
    $}
    \label{eq:zorro_asym}
\end{equation}
\begin{equation}
    \begin{aligned}
        k_{i} = 1 + e^{a_{i} b} \\
        k_{s} = 1 + e^{a_{s} b}
    \end{aligned}
    \label{eq:zorro_const_asym} 
\end{equation}

\subsubsection{Sigmoid-Zorro}
Some architectures, such as LSTM and Variational Autoencoders, use the Sigmoid function. Replacing it with the Zorro function would lead to undesired results due to the difference in the central position and the slope in the linear zone. The Sigmoid and similar functions, such as the Hard-Sigmoid, have a bounded image, varying its values between 0 and 1. In contrast, the Zorro function defined in Equation \ref{ec:zorro} has a bounded image, varying its values between $-k/a$ and $1+k/a$ for $a>0$. So, the \textit{Sigmoid-Zorro} function arises by applying a shift and rescaling as indicated by Equation \ref{eq:zorro_sigmoid}, obtaining a function with behavior more similar to the Sigmoid and allowing its use in the mentioned architectures.
\begin{equation}
    \text{Zorro}_\text{sigmoid}(x, a, b) = \text{Zorro}_\text{sym}((x+2)/4, a, b)
    \label{eq:zorro_sigmoid}
\end{equation}

This variant is symmetric, with a decreasing slope up to $0.25$ and a horizontal offset so that its value at $x=0$ is equal to 0.5, just like the Sigmoid function. Figure \ref{fig:zorro_sigmoid} compares the two functions, showing how this setting produces an activation function with the same central area, principal value, global mean value, and horizontal asymptotics as the Sigmoid function. However, the Zorro variant has the characteristic humps, and its image is broader  than the Sigmoid function. 
This means that even when only a small margin is outside $[0, 1]$, a cumulative application of the Sigmoid will tend to produce small changes (VGP) In contrast, cumulative applications of the Sigmoid-Zorro will tend to explode and go to infinity if the problem conditions are correct.
In this work, we will limit ourselves to showing the behavior of functions on simple architectures and in an application case. The analysis of the particular behavior of Zorro and Sigmoid-Zorro on relevant architectures replacing Sigmoid will be addressed in future work.


\subsubsection{Tanh-Zorro}
The previous function was designed to replace the Sigmoid from gates in LSTM networks. This type of network also uses the Hyperbolic Tangent (Tanh) as the activation function. Following the same sense, we define the Tanh-Zorro function to mimic the Tanh behavior, defined by Equation \ref{eq:zorro_tanh}. This activation function has a mean activation closer to zero, so, as mentioned above, it will produce less bias in subsequent layers and should provide better training performance. 
\begin{equation}
    \text{Zorro}_\text{tanh}(x, a, b) = 2\ \text{Zorro}_\text{sigmoid}(x, a, b) - 1
    \label{eq:zorro_tanh}
\end{equation}

Figure \ref{fig:zorro_tanh} compares the Tanh and the Tanh-Zorro functions. It is easy to see that the Tanh-Zorro function is centered at (0,0), has a linear part between $-1$ and $1$ with slope 1, and has the same asymptotics as the original function. This characteristic allows the Tanh-Zorro variant to replace Tanh on most architectures, considering that the image is more extensive than $[-1, 1]$. A cumulative application could explode even for a small range, and the gradient would tend to be infinite given the right conditions because the Tanh-Zorro image is outside that range.


\subsubsection{Sloped-Zorro}

Previous approaches have shown that increasing the function's derivative will increase the convergence performance \cite{Roodschild2020}. Therefore, increasing the function's slope should produce faster training and better performance and allow the backpropagation algorithm to train networks with more layers (higher VGP resistance). We then propose the Sloped-Zorro variant with a different slope through Equation \ref{eq:zorro_sloped}, reparametrizing the original Zorro function by introducing the $m$ coefficient. This way, differentiability, boundedness, and other important properties are preserved. The effect of the reparametrization can be seen in Figure \ref{fig:zorro_sloped}.
\begin{equation}
    \newlength\eqmaxwidthzs
    \setlength{\eqmaxwidthzs}{\columnwidth-27pt}
    \resizebox{\ifnum\width<\eqmaxwidthzs \width \else \eqmaxwidthzs\fi}{!}{$%
    \text{Zorro}_\text{sloped}(x, a_s, a_i, b, m) = \text{Zorro}_\text{asym}(mx, a_s, a_i, b)
    $}
    \label{eq:zorro_sloped}
\end{equation}


\subsection{Approximations to other activation functions}
\label{Similarities}

\begin{figure}[htb]
    \centering
    \begin{subfigure}{0.493\columnwidth}
        \centering
        \includegraphics[height=0.83\textwidth]{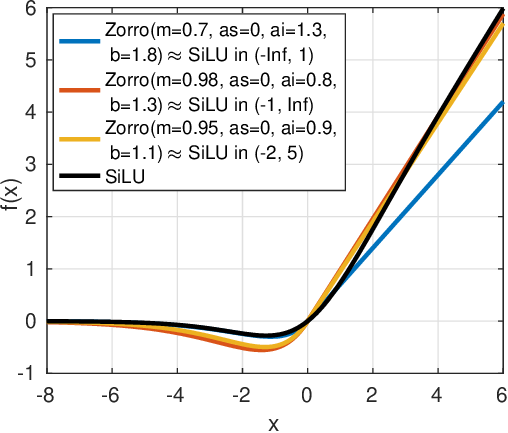}
        \caption{SiLU-Zorro.}
        \label{fig:zorro_silu}
    \end{subfigure}
    \begin{subfigure}{0.493\columnwidth}
    \centering
        \includegraphics[height=0.83\textwidth]{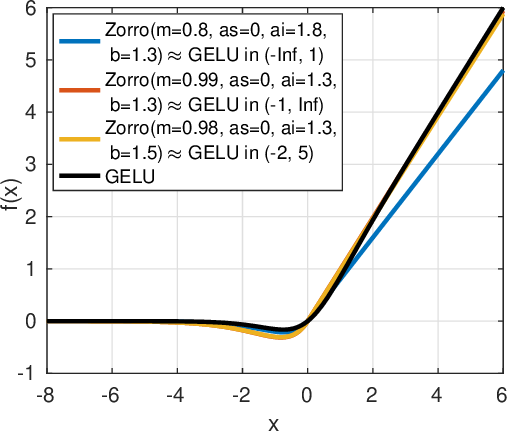}
        \caption{GELU-Zorro.}
        \label{fig:zorro_gelu}
    \end{subfigure}
    \begin{subfigure}{0.493\columnwidth}
        \centering
        \includegraphics[height=0.83\textwidth]{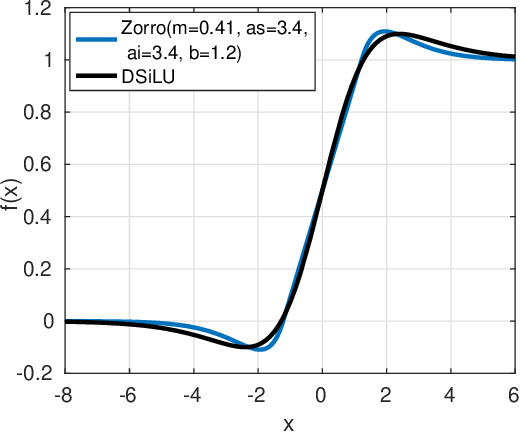}
        \caption{DSiLU-Zorro.}
        \label{fig:zorro_dsilu}
    \end{subfigure}
    \begin{subfigure}{0.493\columnwidth}
        \centering
        \includegraphics[height=0.83\textwidth]{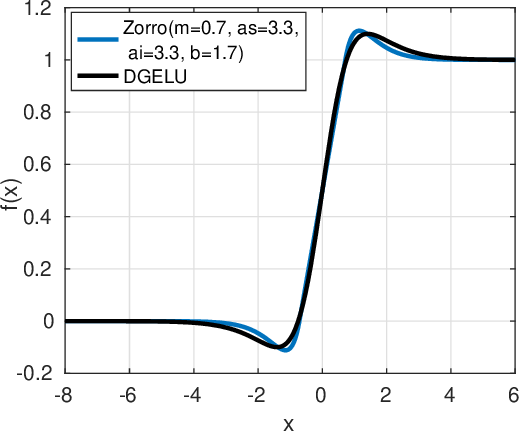}
        \caption{DGELU-Zorro.}
        \label{fig:zorro_dgelu}
    \end{subfigure}
    \begin{subfigure}{0.493\columnwidth}
        \centering
        \includegraphics[height=0.83\textwidth]{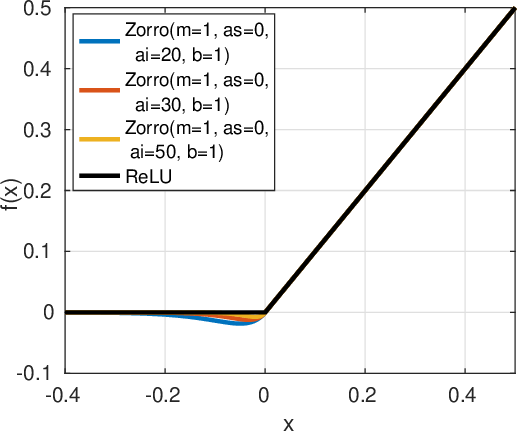}
        \caption{ReLU-Zorro.}
        \label{fig:zorro_relu}
    \end{subfigure}
    \caption{Approximation of some widely used activation functions using the Zorro function.}
    \label{fig:zorro_approx}
\end{figure}

One of the main design objectives of Zorro is to be flexible and represent other widely used activation functions. First, it is possible to approximate the ReLU function by taking the Sloped-Zorro function with $m=1$, $a_{s}=0$, and $a_{i}$ as high as possible. If we assign infinity to the paŕameter $a_{i}$, the Zorro function mathematically becomes ReLU. However, depending on the implementation and the programming language, the function may yield an invalid or infinite value, so assigning a sufficiently large value is acceptable. Figure \ref{fig:zorro_relu} compares ReLU and the Zorro function for different values of its parameter. We can see that $a_{i}=50$ achieves an excellent approximation.

On the other hand, taking an appropriate value for $a_{i}$ and $b_{i}$ from the asymmetric Sloped-Zorro function defined by Equation \ref{eq:zorro_sloped} will approximate the negative part of the Swish function. For positive values, Zorro can only be a straight line or a concave function that eventually becomes asymptotic at $y=1$. Therefore, there is a maximal interval in which variants of Zorro can approximate Swish-based functions and their derivatives due to their general form. Table \ref{tab:approximation_params} shows the different sets of parameters that can be used with Zorro to approximate these functions. The last column shows the maximum difference ($\varepsilon$) between the two functions. 
In the case of SiLU and GELU, several sets of parameters allow Sloped-Zorro to approximate them depending on the input range adopted (see Table \ref{tab:approximation_params}). If normalized data are used or a batch normalization is included in the training, the data will be mostly between 0 and 1, so the approximation interval $(-\infty, 1)$ is the most accurate. However, any of the other approximation intervals could be used to cover a broader domain. These parameters could even be trainable to find a value that best fits each data set.
Figures \ref{fig:zorro_silu} and \ref{fig:zorro_gelu} show the approximation mentioned above. It is clear that, despite having very similar functions, their behavior around zero is very different, and the linear part of ReLU preserved in Zorro makes for a function different from GELU and other variants of Swish.

\begin{table}[htb]
    \centering
    \footnotesize
    \resizebox{\ifnum\width<\columnwidth \width \else \columnwidth \fi}{!}{
    \begin{tabular}{|l|c|cccc|c|}
        \hline
        \multirow{2}{*}{\textbf{\begin{tabular}{@{}c@{}}Activation\\function\end{tabular}}} & 
        \multirow{2}{*}{\textbf{\begin{tabular}{@{}c@{}}Approximation\\interval\end{tabular}}} & 
        \multicolumn{4}{c|}{\textbf{Best parameter values}} &
        \textbf{Max. Error} \\
        \cline{3-6}
        & & 
        \textbf{m} &
        \textbf{\textit{a}${}_i$} & 
        \textbf{\textit{a}${}_s$} & 
        \textbf{b} & 
        \textbf{$\varepsilon$} \\ 
        \hline
        ReLU                  & $(-\infty, \infty)$ & 1.00 & 50.00 & 0.0 & 1.0 & 0.001  \\
        \hline
        \multirow{3}{4em}{SiLU / Swish} & $(-\infty, 1)$      & 0.70 & 1.30  & 0.0 & 1.8 & 0.041  \\
                              & $(-1, \infty)$      & 0.98 & 0.80  & 0.0 & 1.3 & 0.254  \\
                              & $(-2,5)$            & 0.95 & 0.90  & 0.0 & 1.1 & 0.219  \\
        \hline
        \multirow{3}{*}{GELU} & $(-\infty, 1)$      & 0.80 & 1.80  & 0.0 & 1.3 & 0.054  \\
                              & $(-1, \infty)$      & 0.99 & 1.99  & 0.0 & 1.3 & 0.155  \\
                              & $(-2,5)$            & 0.98 & 1.30  & 0.0 & 1.5 & 0.147  \\
        \hline
        DSiLU                 & $(-\infty, \infty)$ & 0.41 & 3.40  & 3.4 & 1.2 & 0.037  \\
        \hline
        DGELU                 & $(-\infty, \infty)$ & 0.70 & 3.30  & 3.3 & 1.7 & 0.036  \\
        \hline
    \end{tabular}
    }
    \caption{Different parameter values of the Sloped-Zorro function to approximate the more commonly used activation functions.}
    \label{tab:approximation_params}
\end{table}

DSiLU and DGELU, the derivatives of the functions SiLU and GELU, respectively, can be approximated by the original Symmetric-Zorro very effectively using the parameters given in Table \ref{tab:approximation_params}. Our tests show, nevertheless, that the linear part original of the ReLU still produces better results than GELU and DGELU in many cases. A thorough comparison of an appropriate case will be conducted in a subsequent paper. Once again, the flexibility of our family of functions will allow for many adaptations to particular architectures and datasets, with only a few parameters during training.

\section{Parameter adjustment}
\label{sec:ajuste_parametros}

Our functions are mainly reparametrizations of each other. Therefore, it is very important to establish a convenient set of parameters for each variant so that future users do not have to fit the parameters to a particular architecture and dataset forcibly. However, we also consider it important to show that these functions exhibit very different behaviors, making them suitable for different architectures and environments.

In order to do that, we use a dense feedforward architecture without batch normalization or convolutions, a fixed training and validation division, and a fixed number of epochs. The choice of a dense feedforward architecture is deliberate: Parameter exploration and activation function response to the vanishing gradient problem makes more sense in feedforward neural networks \cite{pmlr-v9-glorot10a}. Vanishing Gradient, Saturation, and Exploding Gradient problems tend to be more severe and challenging in this kind of network than in more modern architectures such as LSTMs, GRUs, and Transformers, which avoid these issues by design.  These last networks incorporate specific features and channels to respond to these problems more effectively, where deeper layers receive unaltered fresh information. It strongly contrasts ordinary dense feedforward networks, where the information is passed down only from the previous layer.  So, we used a standard feedforward architecture of progressively deeper layers in order to adjust our parameters and study the behavior of the functions. Our experiments showed that, at least in convolutional architectures, the same parameters adjusted in feedforward architectures produced good results without further adjustments.

Now, in order to apply grid search, it is necessary to define an interval and a step value to obtain the parameter sets that we need to evaluate. For this purpose, we consider practical and usable values for training a neural network, discarding extreme values and points where the function is not defined. Table \ref{tab:parameters} contains the intervals and steps used for each parameter of Zorro variants. The conditions of the test are shown in Table \ref{tab:feedforward_architecture}.
\begin{table}[htb]
    \centering
    \footnotesize
    \resizebox{\ifnum\width<\columnwidth \width \else \columnwidth \fi}{!}{
    \begin{tabular}{|l|l|}
        \hline
       \textbf{Parameter} & \textbf{Details} \\
        \hline
            Dataset             & MNIST \\
            Architecture        & Dense Feedforward \\
            Optimizer           & Adam, learning rate: 0.01 \\
            Number of hidden neurons  & 128 \\
            Epochs              & 15   \\
            Batch normalization & No   \\
            Batch Size          & 1024 \\
            Runs                & 4    \\
        \hline
    \end{tabular}
    }
    \caption{Details of the dense Feedforward architecture used for parameter adjustment.}
    \label{tab:feedforward_architecture}
\end{table}

Now, the desirable parameter values are those that allow the training of a deep neural network. So, we progressively increase the depth until the VGP sinks and the network cannot train anymore. We define a parameter set that produces good training as one that obtains more than $90\%$ validation error. Our preliminary tests showed that until VGP occurs, any reasonable set of parameters will produce good training. Then, as the number of layers approaches the VGP point, only a small percentage ($5\%$ to $10\%$) of the parameters will produce good training. We take those limit parameters as the preferred set. 

Then, we define the \textit{Stable Layer} as the maximum number of layers where $40\%$ or more of the considered parameter sets yield good training. In other words, the stable layer is the maximum layer where more than $40\%$ of the parameter sets obtain a validation error of $90\%$. After that layer, the training becomes unstable, and progressively fewer parameter sets achieve the desired validation error. Then, we define the \textit{Maximum Layer} as the final layer that can be successfully and consistently trained (with a smaller percentage of tested parameter sets). Due to the VGP, no parameters can train the network for more layers. For the analysis, the training was repeated 4 times for each parameter set and each number of layers. This number was adopted because preliminary experiments showed they are sufficient to determine whether the training is stable. If the training algorithm is stable for 4 runs, it will train the network successfully when considering more repetitions. If the algorithm fails in these runs, the reason is the VGP or a problem with the architecture used.

Table \ref{tab:capas_logradas} shows the Stable and Maximum layers for each activation function. It reports the number of layers achieved, the percentage of parameter sets that achieve a validation error equal to or greater than $90\%$, and the average training and testing accuracy over the 4 runs performed. The Symmetric-Zorro function successfully trains up to 30 layers with an accuracy of $96.7\%$, and only $52.4\%$ of the parameter sets (around half) get more than $90\%$ validation error. The maximum number of layers is 34, with only $23.8\%$ of the parameter sets training successfully. The accuracy is $96.5\%$, similar to the previous case, indicating a very small decrease in performance for the stable set of parameters chosen.
The other functions present very different stable and maximal layers, ranging from only 15 for Sigmoid-Zorro to 40 for Tanh-Zorro. There are very small differences in the validation error between the stable layer and the maximal layer (less than $1\%$ difference), showing that the training is consistent and successful, or VGP takes place in a noticeable fashion, and the training becomes unstable. Some functions are sensitive and require more parameter adjustment, like Asymmetric-Zorro with $41.7\%$ of successful parameters and Sloped-Zorro with $50.0\%$. Meanwhile, Tanh-Zorro is almost insensitive, and $95.4\%$ of the parameters can efficiently train the network with 38 layers, decreasing to only $12.7\%$ in $43$ layers (meaning that as we approach more complex problems and deeper networks, the need for parameter adjustment might appear). It is also important to notice that the functions and network might eventually train for more layers. However, since that training is not consistently successful, we do not include those results in the table.

\begin{table}[htb]
    \centering
    \footnotesize
    \resizebox{\ifnum\width<\columnwidth \width \else \columnwidth \fi}{!}{
    \begin{tabular}{|l|c@{\ }|c|c@{\ }|c|c@{\ }|c|c@{\ }|c|}
        \hline
        \multirow{2}{*}{\textbf{\begin{tabular}{@{}c@{}}Activation\\function\end{tabular}}} &
        \multicolumn{2}{c|}{\textbf{\begin{tabular}{@{}c@{}}Number of \\layers\end{tabular}}} &
        \multicolumn{2}{c|}{\textbf{\begin{tabular}{@{}c@{}}Percentage with \\validation $>$90\%\end{tabular}}} &
        \multicolumn{2}{c|}{\textbf{\begin{tabular}{@{}c@{}}Max training \\precision [\%]\end{tabular}}} &
        \multicolumn{2}{c|}{\textbf{\begin{tabular}{@{}c@{}}Max validation \\precision [\%]\end{tabular}}} \\
        \cline{2-9}
        & \textbf{\scriptsize Stable} & \textbf{\scriptsize Maximal} & \textbf{\scriptsize Stable} & \textbf{\scriptsize Maximal} & \textbf{\scriptsize Stable} & \textbf{\scriptsize Maximal} & \textbf{\scriptsize Stable} & \textbf{\scriptsize Maximal} \\
        \hline
        \pbox[c][2.5em][c]{5em}{Symmetric-Zorro}
        & 30 & 34 & 52.4 & 23.81 & 98.40 & 97.70 & 96.78 & 96.54 \\
        \hline
        \pbox[c][2.5em][c]{5em}{Asymmetric-Zorro}
        & 40 & 43 & 41.7 & 6.67 & 97.24 & 96.81 & 96.31 & 95.71 \\
        \hline
        \pbox[c][2.5em][c]{5em}{Sigmoid-Zorro}
        & 15 & 16 & 68.0 & 13.33 & 95.25 & 93.38 & 93.94 & 92.94 \\
        \hline
        \pbox[c][2.5em][c]{5em}{Tanh-Zorro}
        & 38 & 43 & 95.4 & 12.73 & 97.07 & 96.78 & 96.07 & 95.70 \\ 
        \hline
        \pbox[c][2.5em][c]{5em}{Sloped-Zorro}
        & 23 & 30 & 50.0 & 11.67 & 98.69 & 98.06 & 96.82 & 96.58 \\
        \hline
    \end{tabular}
    }
    \caption{Parameter adjustment for the new activation functions - Stable and Maximal layer for each function.}
    \label{tab:capas_logradas}
\end{table}

\begin{figure*}[!tb]
    \centering
    \begin{subfigure}{0.193\textwidth}
        \centering
        \includegraphics[height=0.80\textwidth]{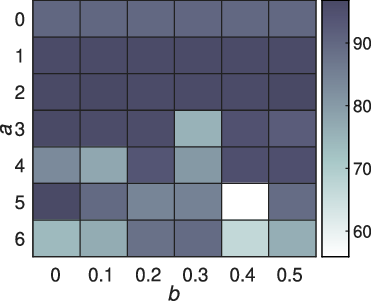}
        \caption{Symmetric-Zorro(Stable).}
        \label{fig:_parametros_zorro_original_1}
    \end{subfigure}
    \begin{subfigure}{0.193\textwidth}
        \centering
        \includegraphics[height=0.80\textwidth]{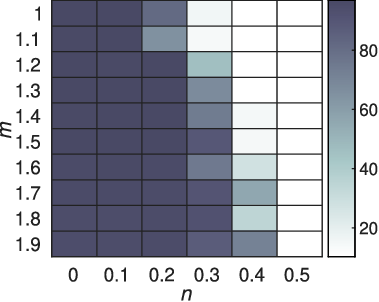}
        \caption{Sloped-Zorro(Stable).}
        \label{fig:parametros_zorro_sloped_1}
    \end{subfigure}
    \begin{subfigure}{0.193\textwidth}
        \centering
        \includegraphics[height=0.80\textwidth]{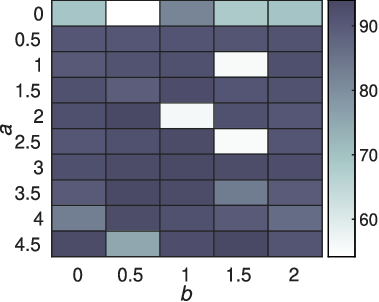}
        \caption{Sigmoid-Zorro(Stable).}
        \label{fig:parametros_zorro_sigmoid_1}
    \end{subfigure}
    \begin{subfigure}{0.193\textwidth}
        \centering
        \includegraphics[height=0.80\textwidth]{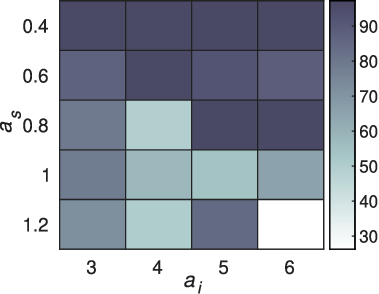}
        \caption{Asymmetric-Zorro(Stable).}
        \label{fig:parametros_zorro_asimetrica_1}
    \end{subfigure}
    \begin{subfigure}{0.193\textwidth}
        \centering
        \includegraphics[height=0.80\textwidth]{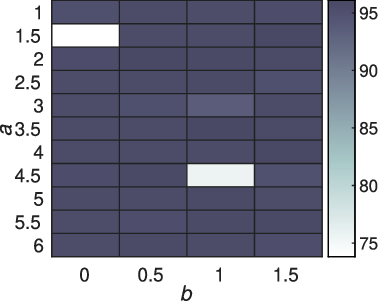}
        \caption{Tanh-Zorro(Stable).}
        \label{fig:parametros_zorro_tanh_1}
    \end{subfigure}
    \begin{subfigure}{0.193\textwidth}
        \centering
        \includegraphics[height=0.80\textwidth]{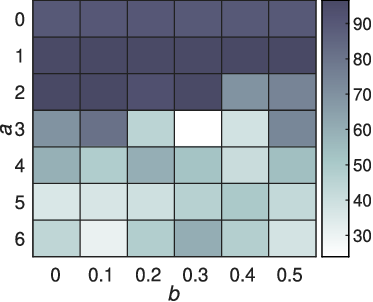}
        \caption{Symmetric-Zorro(Maximal).}
        \label{fig:parametros_zorro_original_2}
    \end{subfigure}
    \begin{subfigure}{0.193\textwidth}
        \centering
        \includegraphics[height=0.80\textwidth]{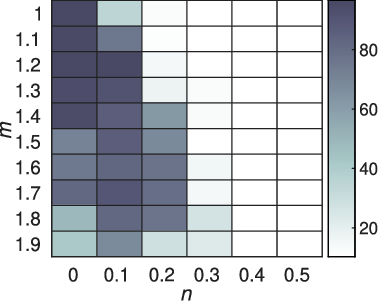}
        \caption{Symmetric-Zorro(Maximal).}
        \label{fig:parametros_zorro_sloped_2}
    \end{subfigure} 
    \begin{subfigure}{0.193\textwidth}
        \centering
        \includegraphics[height=0.80\textwidth]{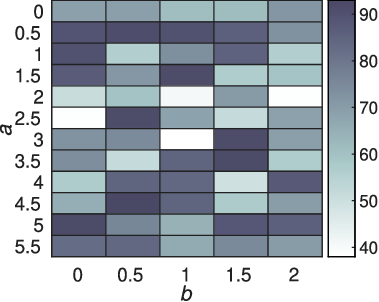}
        \caption{Sigmoid-Zorro(Maximal).}
        \label{fig:parametros_zorro_sigmoid_2}
    \end{subfigure}
    \begin{subfigure}{0.193\textwidth}
        \centering
        \includegraphics[height=0.80\textwidth]{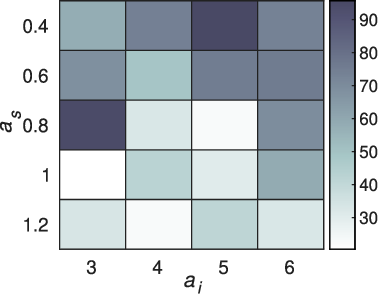}
        \caption{Asym-Zorro(Maximal).}
        \label{fig:parametros_zorro_asimetrica_2}
    \end{subfigure}
    \begin{subfigure}{0.193\textwidth}
        \centering
        \includegraphics[height=0.80\textwidth]{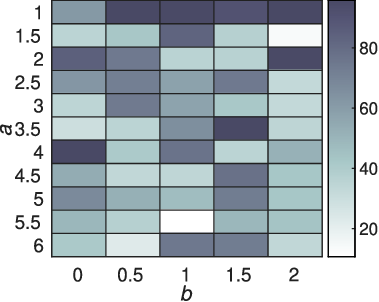}
        \caption{Tanh-Zorro(Maximal).}
        \label{fig:parametros_zorro_tanh_2}
    \end{subfigure}
    \caption{Validation error surfaces for the stable and maximum layers obtained for the different parameters of the Zorro function.}
    \label{fig:parametros_capas}
\end{figure*}

Figure \ref{fig:parametros_capas} shows the validation errors achieved using each of the parameter values considered, corresponding to the Stable and Maximal layers, respectively. It can be clearly observed that the stable layers have a higher percentage of successful training (darker squares). Using more layers does not necessarily mean failed training, only that the training will be more unstable. Finally, Table \ref{tab:parameters} shows the parameters that produce the best results in the Stable and Maximal layers. The parameters that produce the best result in the Maximum layer are not necessarily stable for different runs. So, this table includes the set of parameters that produce stable results in the Maximum layer, although they are not necessarily the most accurate.

\begin{table}[htb]
\centering
\footnotesize
\resizebox{\ifnum\width<\columnwidth \width \else \columnwidth \fi}{!}{
\begin{tabular}{|l|c|c|c|c|c|}
\hline
  \multirow{2}{*}{\textbf{\begin{tabular}{@{}c@{}}Activation\\function\end{tabular}}} &
  \multirow{2}{*}{\textbf{\begin{tabular}{@{}c@{}}Parameter\\interval\end{tabular}}} &
  \multirow{2}{*}{\textbf{\begin{tabular}{@{}c@{}}Parameter\\step\end{tabular}}} &
  \multicolumn{1}{c|}{\textbf{\begin{tabular}{@{}c@{}}Stable Layer\end{tabular}}} &
  \multicolumn{2}{c|}{\textbf{\begin{tabular}{@{}c@{}}Maximal Layer\end{tabular}}} \\ 
 \cline{4-6} 
 & & & \textbf{\scriptsize Best value} & \textbf{\scriptsize Best results} & \textbf{\scriptsize Stable results} \\ \hline
\multirow{2}{5em}{Symmetric-Zorro} 
    & $a \in [0, 6]$   & 1.0 & 2.0 & 1.0 & 2.0 \\
    & $b \in [0, 0.5]$ & 0.1 & 0.5 & 0.5 & 0.3 \\ \hline
\multirow{3}{5em}{Asymmetric-Zorro} 
    & $a_{i} \in [3, 6]$     & 1.0 & 6.0 & 5.0 & 5.0 \\
    & $a_{s} \in [0.4, 1.2]$ & 0.2 & 0.8 & 0.4 & 0.4 \\ 
    & $b \in [0, 0.4]$       & 0.2 & 0.4 & 0.4 & 0.4 \\ \hline
\multirow{2}{5em}{Sigmoid-Zorro} 
    & $a \in [0, 5.5]$ & 0.5 & 2.0 & 4.5 & 3.0 \\ 
    & $b \in [0, 2]$   & 0.5 & 0.5 & 0.5 & 1.5 \\ \hline
\multirow{2}{5em}{Tanh-Zorro} 
    & $a \in [1, 6]$   & 0.5 & 3.5 & 4.0 & 3.5 \\ 
    & $b \in [0, 1.5]$ & 0.5 & 1.0 & 1.0 & 1.5 \\ \hline
\multirow{4}{5em}{Sloped-Zorro} 
    & $a \in [0, 6]$   & 1.0 & 2.0 & 2.0 & 2.0 \\
    & $b \in [0, 6]$   & 0.1 & 0.3 & 0.3 & 0.3 \\ 
    & $m \in [1, 2]$   & 0.1 & 1.3 & 1.2 & 1.2 \\
    & $n \in [0, 0.5]$ & 0.1 & 0.0 & 0.1 & 0.0 \\ \hline
\end{tabular}
}
\caption{Parameters for each activation function - Best values for stable and maximal layer, and most frequent value in the maximal layer.}
\label{tab:parameters}
\end{table}


\section{Application to Convolutional Networks}
\label{sec:cnn}

The experiments performed in the previous section used feedforward networks with fully connected layers without applying convolution. We proceeded with this method to study the impact on the Vanishing Gradient Problem and highlight the differences between Zorro variants as activation functions. In this section, we will apply the variants of Zorro to simple convolutional networks on known datasets to show the improvements that can be achieved. Convolutional Neural Networks (CNNs) \cite{huang2018densely} are a class of feedforward networks specialized in processing input data with a grid-like topology, typically images. CNNs exploit images' spatial structure and features to extract and learn relevant features. CNNs are crucial in the fields of deep learning and computer vision. Important and widespread libraries like Keras, PyTorch, and OpenCV include pre-trained Convolutional architectures VGG, ResNET, Inception, and YOLO (You Only Look Once)  \cite{agronomy13061633}. This last one is a leading library in the world for computer vision. Improving these networks leads to better performance in computer vision tasks, which is essential in practical applications such as security systems, medical diagnostics, autonomous vehicles, and more. 

A CNN takes an input image and automatically assigns importance to various aspects or features of the image through adjustable weights and biases, effectively differentiating between different elements during training. The architecture is based on convolutional, pooling, and fully connected layers. The convolutional layer is the core of the CNN. It performs most of the computational work necessary for feature extraction using filters. The pooling layer reduces the spatial size of the input and extracts the most dominant information without losing important image properties. This pooling layer reduces processing needs, acts as a noise suppressor, and maintains translational invariance. It means that if the input image is shifted, the output of the pooling layer does not change significantly. At the end of the convolutional and pooling layers, fully connected layers are typically used for the final task.  

We will use simple convolutional neural networks to test the new Zorro activation functions against more traditional ones. Comparisons of GELU and DGELU with their respective approximations using the Zorro function are of particular interest. The selected databases are some of the most widely used in the machine learning literature: MNIST \cite{deng2012mnist}, Fashion MNIST \cite{xiao2017fashion}, CIFAR-10, Letters EMNIST \cite{cohen2017emnist}, and Balanced EMNIST \cite{cohen2017emnist}. The architecture used for CNN is described in Table \ref{tab:cnn_architecture}. Each convolutional layer was followed by the activation function under observation. Although these are small architectures, they are sufficient to demonstrate the validity of the approach and evaluate the impact of the proposed activation functions on CNN performance.
Hypothesis tests were performed to compare the results and determine whether there are significant differences in the means of the precisions achieved in the different runs corresponding to the different activation functions used. Specifically, we use Welch's t-test \cite{Welch1947Generalization} for two independent samples, assuming the two population variances are different. We are using the Student's t Distribution, given that 10 runs were performed. 

\begin{table}[htb]
    \centering
    \footnotesize
    \resizebox{\ifnum\width<\columnwidth \width \else \columnwidth \fi}{!}{
    \begin{tabular}{|l|l|}
        \hline
       \textbf{Parameter} & \textbf{Details} \\
        \hline
        Input shape          & $28\times28\times1$ for single-channel images \\
        Normalization        & No \\
        Batch size           & 1024 for Letters and Balanced databases \\
                              & 2048 for other databases \\
        Optimizer            & Adam, SGD (with momentum), learning rate: 0.001 \\
        Loss function        & Categorical Cross entropy \\
        Training epochs	  & 30 \\
        \hline
        \multirow{1}{*}{Layers}
                             & Convolution 2D (filters: 4, kernel size: $3\times3$) \\
                             & Convolution 2D (filters: 4, kernel size: $3\times3$) \\
                             & Max Pooling 2D (size: $2\times2$) \\
                             & Dropout (0.25) \\
                             & Flatten\\
                             & Dense Feedforward (512 units) \\
                             & Dropout (0.5) \\
                             & Output (activation: softmax) \\
        \hline
    \end{tabular}
    }
    \caption{Details of the Convolutional Neural Network (CNN) architecture used to test different activation functions.}
    \label{tab:cnn_architecture}
\end{table}

\begin{table*}[tb]
    \centering
    \footnotesize
    \resizebox{\ifnum\width<\textwidth \width \else \textwidth \fi}{!}{
    \begin{tabular}{|l|c@{\ }c@{\ }c@{\ }c|c@{\ }c@{\ }c@{\ }c|c@{\ }c@{\ }c@{\ }c|c@{\ }c@{\ }c@{\ }c|c@{\ }c@{\ }c@{\ }c|}
    \hline
        \textbf{Activation} & 
        \multicolumn{4}{c|}{\textbf{CIFAR-10}} & 
        \multicolumn{4}{c|}{\textbf{Fashion MNIST}} & 
        \multicolumn{4}{c|}{\textbf{MNIST}} & 
        \multicolumn{4}{c|}{\textbf{Letters EMNIST}} & 
        \multicolumn{4}{c|}{\textbf{Balanced EMNIST}} \\
    \cline{2-21}
        \textbf{Function} & 
        \textbf{Max.} & \textbf{Mean} & \textbf{STD} & \textbf{p} & 
        \textbf{Max.} & \textbf{Mean} & \textbf{STD} & \textbf{p} & 
        \textbf{Max.} & \textbf{Mean} & \textbf{STD} & \textbf{p} & 
        \textbf{Max.} & \textbf{Mean} & \textbf{STD} & \textbf{p} & 
        \textbf{Max.} & \textbf{Mean} & \textbf{STD} & \textbf{p} \\
    \hline
        Sigmoid & 
            43.13 &          40.39 & 1.95 & \textbf{0.00} & 
            83.09 &          81.97 & 0.58 & \textbf{0.00} & 
            94.89 &          94.51 & 0.25 & \textbf{0.00} & 
            3.85  &          3.85  & 0.00 & \textbf{0.00} & 
            2.13  &          2.13  & 0.00 & \textbf{0.00}\\
        Tanh    & 
            54.90 &          51.91 & 1.77 & \textbf{0.00} & 
            88.61 &          88.33 & 0.22 & \textbf{0.00} & 
            98.34 &          97.91 & 0.26 & \textbf{0.00} & 
            83.38 &          80.96 & 1.51 & \textbf{0.00} & 
            76.38 &          73.51 & 2.15 & \textbf{0.00}\\
        ReLU    & 
            61.10 & \textbf{59.13} & 1.32 &          1.00 & 
            90.63 & \textbf{90.03} & 0.32 &          1.00 & 
            99.00 & \textbf{98.76} & 0.20 &          1.00 & 
            90.44 & \textbf{88.78} & 2.07 &          1.00 & 
            83.09 & \textbf{82.52} & 0.36 &          1.00\\
        ELU     & 
            55.21 &          53.00 & 1.41 & \textbf{0.00} & 
            88.95 &          88.68 & 0.20 & \textbf{0.00} & 
            98.65 &          98.52 & 0.10 & \textbf{0.00} & 
            89.04 &          88.54 & 0.34 &          0.73 & 
            80.51 &          79.05 & 1.41 & \textbf{0.00}\\
        SiLU    & 
            57.27 &          56.23 & 1.06 & \textbf{0.00} & 
            89.16 &          88.78 & 0.24 & \textbf{0.00} & 
            98.93 &          98.67 & 0.17 &          0.25 & 
            90.45 & \textbf{90.24} & 0.14 &          0.05 & 
            83.35 & \textbf{83.20} & 0.13 & \textbf{0.00}\\
        GELU    & 
            60.02 &          58.24 & 1.21 &          0.13 & 
            90.11 &          89.64 & 0.26 & \textbf{0.01} & 
            98.94 &          98.73 & 0.09 &          0.68 & 
            90.63 & \textbf{90.37} & 0.12 & \textbf{0.04} & 
            83.87 & \textbf{83.70} & 0.15 & \textbf{0.00}\\
        DGELU   & 
            55.53 &          54.08 & 0.90 & \textbf{0.00} & 
            88.77 &          88.28 & 0.25 & \textbf{0.00} & 
            98.29 &          98.05 & 0.23 & \textbf{0.00} & 
            89.27 &          46.22 & 44.67& \textbf{0.02} & 
            2.13  &          2.13  & 0.00 & \textbf{0.00} \\
    \hline        
        $\text{Zorro}_\text{sigmoid}$ & 
        47.70 &          45.90 & 1.04 & \textbf{0.00} & 
        84.61 &          83.81 & 0.50 & \textbf{0.00} & 
        96.43 &          95.78 & 0.45 & \textbf{0.00} & 
        67.03 &          59.17 & 19.49& \textbf{0.00} & 
        65.01 &          61.55 & 1.92 & \textbf{0.00} \\
        $\text{Zorro}_\text{tanh}$    & 
        53.05 &          51.59 & 1.30 & \textbf{0.00} & 
        87.75 &          87.60 & 0.13 & \textbf{0.00} & 
        98.24 &          97.58 & 0.31 & \textbf{0.01} & 
        83.18 &          79.50 & 3.24 & \textbf{0.00} & 
        78.10 &          72.94 & 2.04 & \textbf{0.00} \\
        $\text{Zorro}_\text{sym}$     & 
        62.97 & \textbf{60.26} & 1.41 &          0.08 & 
        90.96 & \textbf{90.62} & 0.30 & \textbf{0.00} & 
        99.06 & \textbf{98.90} & 0.09 &          0.06 & 
        90.98 & \textbf{90.86} & 0.11 & \textbf{0.01} & 
        84.70 & \textbf{84.43} & 0.14 & \textbf{0.00} \\
        $\text{Zorro}_\text{asym}$    & 
        62.37 &          58.51 & 2.65 &          0.52 & 
        90.96 & \textbf{90.51} & 0.27 & \textbf{0.00} & 
        98.94 & \textbf{98.85} & 0.07 &          0.22 & 
        91.04 & \textbf{90.31} & 0.35 & \textbf{0.05} & 
        84.60 & \textbf{83.94} & 0.29 & \textbf{0.00} \\
        $\text{Zorro}_\text{sloped}$  & 
        63.36 & \textbf{61.51} & 1.25 & \textbf{0.00} & 
        91.40 & \textbf{90.93} & 0.24 & \textbf{0.00} & 
        99.01 & \textbf{98.93} & 0.08 & \textbf{0.02} & 
        91.63 & \textbf{91.11} & 0.20 & \textbf{0.01} & 
        85.44 & \textbf{84.82} & 0.30 & \textbf{0.00} \\
    \hline
        $\text{Zorro}_\text{relu}$    & 
        60.38 &          58.28 & 1.44 &          0.19 & 
        90.61 &          90.00 & 0.50 &          0.88 & 
        98.90 &          98.71 & 0.15 &          0.49 & 
        90.38 & \textbf{89.97} & 0.27 &          0.11 & 
        84.47 & \textbf{83.45} & 0.57 & \textbf{0.00} \\
        $\text{Zorro}_\text{gelu1}$   & 
        60.84 &          58.77 & 1.17 &          0.52 & 
        90.60 & \textbf{90.14} & 0.34 &          0.46 & 
        98.95 &          98.72 & 0.19 &          0.61 & 
        91.02 & \textbf{90.76} & 0.18 & \textbf{0.02} & 
        84.85 & \textbf{84.13} & 0.30 & \textbf{0.00} \\
        $\text{Zorro}_\text{gelu2}$   & 
        61.41 &          59.05 & 2.42 &          0.93 & 
        91.23 & \textbf{90.66} & 0.27 & \textbf{0.00} & 
        99.07 & \textbf{98.90} & 0.08 & \textbf{0.04} & 
        91.68 & \textbf{90.98} & 0.27 & \textbf{0.01} & 
        84.61 & \textbf{84.44} & 0.10 & \textbf{0.00} \\
        $\text{Zorro}_\text{gelu3}$   & 
        61.70 & \textbf{59.23} & 1.54 &          0.88 & 
        90.62 & \textbf{90.06} & 0.32 &          0.79 & 
        99.05 & \textbf{98.87} & 0.14 &          0.17 & 
        91.18 & \textbf{90.79} & 0.25 & \textbf{0.01} &
        85.04 & \textbf{84.44} & 0.24 & \textbf{0.00} \\
        $\text{Zorro}_\text{dgelu}$   & 
        56.41 &          53.57 & 1.90 & \textbf{0.00} & 
        88.83 &          88.58 & 0.22 & \textbf{0.00} & 
        98.42 &          98.05 & 0.25 & \textbf{0.00} & 
        88.70 &          20.79 & 35.73& \textbf{0.00} & 
        2.13  &          2.13  & 0.00 & \textbf{0.00} \\
    \hline
    \end{tabular}
    }
    \caption{Results for the different tested activation functions on a Convolutional Architecture for 10 repetitions.}
    \label{tab:CNNPruebas}
\end{table*}

Table \ref{tab:CNNPruebas} shows the accuracy of the validation set obtained by each activation function on each database. They were grouped into four parts: the first contains the traditional functions, including Sigmoid, ReLU, and Swish-based; the second corresponds to DGELU, separated from the others because it has not been proposed as an activation function in previous papers; the third contains the Zorro variants proposed in this work; and the last part includes the Zorro function with parameter setting that approximate the GELU and DGELU function. The average accuracy values exceeding the accuracy obtained by the ReLU function are highlighted in bold style.
It is important to note that the accuracy reported for CIFAR-10 is significantly low compared to state-of-the-art results, which achieve $95\%$ accuracy because we do not use Fractional Max Pooling \cite{Pag_Papers_with_Code_Activ}. However, this simple architecture is sufficient to demonstrate the efficiency of our approach. 
The fourth column of each database shows the p-value of the statistical test that allows deciding when the average accuracy for each activation function differs significantly from the average accuracy of the ReLU function at a significance level $\alpha=0.05$. If the p-value is less than $\alpha$ the hypothesis can be rejected, and the alternate hypothesis can be accepted (the mean values are different). That is, the mean accuracy from the analyzed activation function is statistically different from the ReLU mean accuracy (these cases are indicated in a bold style).

The best function is Sloped-Zorro, which gives the best average accuracy for all databases: $61.51\%$ for CIFAR-10, $90.93\%$ for Fashion MNIST, $98.93\%$ for MINST, $91.11\%$ for Letters, and $84.82\%$ for Balanced Letters. ReLU gives better results for traditional functions for CIFAR-10, Fashion MNIST, and MNIST, whereas GELU is the best for Letters and Balanced Letters. GELU is better than SiLU, but the difference is very small. The Sigmoid function generates a higher error, causing the network not to train for the Letters and Balanced Letters datasets. On the other hand, DGELU is one of the worst in the group of traditional functions, with clearly unstable behavior in the last two datasets (high standard deviation value for one case and validation accuracy close to zero in the other case).

The results for Zorro (see the second group of rows in Table \ref{tab:CNNPruebas}) show that the Symmetric, Asymmetric, and Sloped variants are the best compared to Sigmoid-Zorro and Tanh-Zorro. They are most clearly seen in the average errors reported for the last two data sets. This behavior is possibly due to the linear part of the Zorro function being located in the positive part for the first variants. In contrast, the linear part is centered at 0 for the Sigmoid and Tanh variants of Zorro. However, the Zorro-Sigmoid and variant Zorro-Tanh functions perform better than the original Sigmoid and Tanh functions. 
Also, Symmetric, Asymmetric, and Sloped variants obtain better results than ReLU in most cases, and we can state that the mean values are statistically different. 

On the other hand, the approximating functions of ReLU and DGELU are among the best functions tested. If we compare the original functions with their respective approximations, we see that the Zorro-based versions are better on average, although the differences are very small. Table \ref{tab:cnn-p-values} shows the results of the hypothesis test where the average accuracy of the activation function and the average accuracy of the corresponding approximation to the Zorro function do not present significant differences. It can be observed that the hypothesis cannot be rejected for ReLU and ReLU-Zorro, so the means are equal. It is an expected behavior since the ReLU approximation is very accurate, as seen in Section \ref{Similarities}. Similar results were obtained for DGELU and DGELU-Zorro, except for the Fashion MNIST dataset, where Zorro obtained a slight improvement of 88.6\% over 88.3\%. Finally, for GELU and its approximations, the hypothesis is rejected in all cases except MNIST because the functions are not as close as in the previous cases. However, the accuracy obtained with Zorro is slightly higher than the original function.

\begin{table}[htb]
    \centering
    \footnotesize
    \resizebox{\ifnum\width<\columnwidth \width \else \columnwidth \fi}{!}{
    \begin{tabular}{|l|ccccc|}
    \hline
    \textbf{Function} & \textbf{CIFAR-10} & \textbf{Fashion MNIST} & \textbf{MNIST} & \textbf{Letters} & \textbf{Balanced} \\
    \hline
    $\text{Zorro}_\text{relu}$  & 0.188    & 0.883          &          0.500 & 0.105          & \textbf{0.001} \\
    $\text{Zorro}_\text{gelu1}$ & 0.338    & \textbf{0.000} &          0.803 & \textbf{0.000} & \textbf{0.001}  \\
    $\text{Zorro}_\text{gelu2}$ & 0.362    & \textbf{0.000} & \textbf{0.000} & \textbf{0.000} & \textbf{0.000}  \\
    $\text{Zorro}_\text{gelu3}$ & 0.127    & \textbf{0.000} & \textbf{0.021} & \textbf{0.000} & \textbf{0.000}  \\
    $\text{Zorro}_\text{dgelu}$ & 0.456    & \textbf{0.000} &          0.985 & 0.178          & 1.000           \\
    \hline
    \end{tabular}
    }
    \caption{p-values obtained from hypothesis test for means comparison for ReLU, GELU, and DGELU and their respective Zorro approximations using the CNN architecture.}
    \label{tab:cnn-p-values}
\end{table}


\section{Application to Transform Neural Network}
\label{sec:transformer}

We followed an example using the CIFAR-100 database and 8-layered Transformer architecture taken from \cite{Transformer_2021}, version for Keras 2, modified on 10-01-2022. The architecture was trained from scratch in a GPU following the details contained in Table \ref{tab:transformer_architecture}.

\begin{table}[htb]
    \centering
    \footnotesize
    \resizebox{\ifnum\width<\columnwidth \width \else \columnwidth \fi}{!}{
    \begin{tabular}{|l|l|}
        \hline
        \textbf{Parameter} & \textbf{Details} \\
        \hline
        Input shape        &  $32 \times 32 \times 3$ for three-channel images  \\
        Data augmentation  & Random transformations (output shape: 72$\times$72$\times$3) \\
        Total parameters   & Trainable: 21,787,755, Non-Trainable: 7 \\
        Optimizer          & AdamW, weight decay 0.0001, learning rate 0.001 \\
        Loss function      & Sparse Categorical Cross-entropy  \\
        Training Epochs    & 70 \\
        \hline
        \multirow{1}{*}{Layers}
            & Shifted Patch Tokenization (output shape: 12$\times$12$\times$540) \\
            & Patch Encoder \\
            & \textbf{Transformer Blocks (repeated 8 times):} \\
            & \quad - Layer Normalization (if indicated)\\
            & \quad - Multi-Head with Attention Local Self Attention \\
            & \quad - Add (residual connection) \\
            & \quad - Layer Normalization (if indicated)\\
            & \quad - Dense (128 units), custom Activation Function\\
            & \quad - Dropout (0.1) \\
            & \quad - Dense (64 units), custom Activation Function \\
            & \quad - Dropout (0.1) \\
            & \quad - Add (residual connection) \\
            & Layer Normalization (if indicated) \\
            & Flatten \\
            & Dropout (0.5)\\
            & Dense (2048 units), custom Activation Function\\
            & Dropout (0.5)\\
            & Dense (1024 units), custom Activation Function\\
            & Dropout (0.5)\\
            & Output Dense Layer (100 units), custom Activation\\
        \hline
    \end{tabular}
    }
    \caption{Details of the Transformer-based neural network architecture.}
    \label{tab:transformer_architecture}
\end{table}

Our results are shown in Table \ref{tab:Resultados_Transformer}, indicating the maximum and mean values of 10 runs, the standard deviation, and the p-value of the hypothesis test to compare the mean accuracy achieved with each function and their respective approximate Zorro functions. 
It shows that the $\text{Zorro}_\text{relu}$ variant can effectively replace ReLU with no significant difference in accuracy. Meanwhile, the variants $\text{Zorro}_\text{gelu1}$, $\text{Zorro}_\text{gelu2}$, and $\text{Zorro}_\text{gelu3}$ present a very small difference below GELU, but there is no statistical evidence to reject the hypothesis and state that they are different. Therefore, for an approach based on adapting the activation function to a particular dataset, using these functions is a good starting point when the original architecture contains ReLU, GELU, or similar Swish-based functions.

\begin{table}[htb]
    \centering
    \footnotesize
    \resizebox{\ifnum\width<\columnwidth \width \else \columnwidth \fi}{!}{
    \begin{tabular}{|l|c|c|cccc|}       \hline
         \textbf{Activation} & \multirow{2}{*}{\textbf{Normalization}} & \multirow{2}{*}{\textbf{Testing}} & \multicolumn{4}{c|}{\textbf{Top 5 Accuracy}} \\
         \cline{4-7}
         \textbf{functions} & & & \textbf{Max.} & \textbf{Mean} & \textbf{STD} & \textbf{p-value} \\ 
     \hline
        ReLU                         & Yes & 0.566 & 0.835 &          0.831 & 0.002 & -     \\
    \hline
        ReLU                         & No  & 0.568 & 0.839 &          0.834 & 0.004 & -     \\
        $\text{Zorro}_\text{relu}$   & No  & 0.566 & 0.841 &          0.833 & 0.004 & 0.583 \\ 
    \hline
        GELU                         & No  & 0.576 & 0.843 &          0.837 & 0.004 & -     \\
        $\text{Zorro}_\text{gelu1}$  & No  & 0.573 & 0.841 &          0.836 & 0.004 & 0.583 \\
        $\text{Zorro}_\text{gelu2}$  & No  & 0.569 & 0.842 &          0.835 & 0.004 & 0.278 \\
        $\text{Zorro}_\text{gelu3}$  & No  & 0.568 & 0.839 &          0.834 & 0.004 & 0.111 \\
    \hline
        DGELU                       & No  & 0.359  & 0.694 &          0.683 & 0.006 & - \\
        $\text{Zorro}_\text{dgelu}$ & No  & 0.383  & 0.718 & \textbf{0.705} & 0.005 & \textbf{0.000} \\ 
    \hline         
    \end{tabular}
    }
    \caption{Results for the different activation functions on a Transformer Architecture with 8 layer groups for 10 repetitions on the CIFAR-100 dataset.}
    \label{tab:Resultados_Transformer}
\end{table}

Finally, $\text{Zorro}_\text{dgelu}$ improves over DGELU (0.705 vs. 0.683), although these functions achieved the lowest performance. The other Zorro variants described in this work are not shown because they fail to train the network adequately, achieving even lower accuracy values. Improving the accuracy by incorporating changes in the activation function in this architecture and dataset is a challenge. More tests and adjustments are necessary in order to modify the activation in this architecture in a useful way. 


\section{Conclusions}
\label{sec:conclusions}

A set of 5 flexible, fast, and adaptive activation functions has been introduced in this paper. We have demonstrated that our family of parametric functions can approximate ReLU, Swish, SiLU, DSiLU, GELU, and DGELU. This characteristic makes them ideal for studying parametric adaptation in architectures and packages that involve these functions, such as Yolo V8, xLSTM, and many LLMs and Transformers.  We also showed that they assimilate many features of traditional functions, such as Tanh, while retaining the central linear part that has made ReLU successful.
It is important to note that more variants are clearly possible but have not been studied or proposed here: A variant with a fixed height different from 1, a sloped variant whose maximum and minimum change along with the slope, and many others. This work only depicted the most important and promising variants that proved more useful in our test.

For these variants, we provided a parameter adjustment that should provide future users with an initial parameter and behavior that should be considered a criterion to decide which variants should be best for each particular case. Moreover, as shown in the experiments, the new activation functions produce good results in the convolutional networks. More research is necessary to find improvements in Transformer architecture, but the variants we provide can effectively replace ReLU, GELU, and DGELU for a parametric or heuristic search.

The proposed values should give a good starting point for any architecture and dataset. In subsequent works, we will present an in-depth analysis showing the effectiveness of the Zorro variants in other CNNs and LSTMs. Also, we will study the parameters to be trainable and heuristics to avoid lengthy parameter adjustment processes for each case.

\section*{Funding}
The research leading to these results received funding from Universidad Tecnológica Nacional under Grant Agreement No SITCTU10258C related to ``Automatic Hyperparameter Design in Deep Neural Networks''research project.

\bibliographystyle{plain}
{\footnotesize
\bibliography{main}}

\end{document}